\documentclass[conference]{IEEEtran}
\IEEEoverridecommandlockouts
\usepackage{cite}
\usepackage{amsmath,amssymb,amsfonts}
\usepackage{algorithmic}
\usepackage{graphicx}
\usepackage{textcomp}
\usepackage[table]{xcolor}
\graphicspath{{./}}

\usepackage{multirow}
\usepackage{algorithm}
\usepackage[normalem]{ulem}
\usepackage{booktabs}
\usepackage{stfloats}
\usepackage{placeins}
\usepackage{microtype}

\usepackage{url}
\newcommand{\boldres}[1]{\textcolor{red}{\textbf{#1}}}
\newcommand{\secondres}[1]{\textcolor{blue}{#1}}
\def\BibTeX{{\rm B\kern-.05em{\sc i\kern-.025em b}\kern-.08em
    T\kern-.1667em\lower.7ex\hbox{E}\kern-.125emX}}
\begin{document}

\bstctlcite{IEEEexample:BSTcontrol}

\title{Scale-Aware Pretraining of Time Series\\Foundation Models via Multi-Patch\\Token Alignment and Hybrid Masking
}

\author{
\IEEEauthorblockN{
Taihua Chen\textsuperscript{1,2},
Xiang Ma\textsuperscript{1},
Yixin Zhang\textsuperscript{3},
Tailin Zhan\textsuperscript{1},\\
Manyu Sun\textsuperscript{1}, and
Lizhen Cui\textsuperscript{1,2,*}
}
\IEEEauthorblockA{
\textsuperscript{1}School of Software, Shandong University, Jinan 250101, China\\
\textsuperscript{2}Joint SDU--NTU Centre for Artificial Intelligence Research (C-FAIR),\\
Shandong University, Jinan 250101, China\\
\textsuperscript{3}Nanyang Technological University, Singapore\\[2pt]
\texttt{cfair-cth@mail.sdu.edu.cn},
\texttt{xiangma@sdu.edu.cn},
\texttt{zhangyixin9610@gmail.com},\\
\texttt{tailinzhan@mail.sdu.edu.cn},
\texttt{sjxyctcsusud6383@gmail.com},
\texttt{clz@sdu.edu.cn}
}
\thanks{\textsuperscript{*}Corresponding author: Lizhen Cui (clz@sdu.edu.cn).}
}

\maketitle

\begin{abstract}
Pretraining time series foundation models across heterogeneous datasets necessitates effective handling of varying sampling frequencies. Current methods either employ dataset-specific patch sizes and separate FFNs, leading to fragmented representations, or enforce a fixed patch size that neglects inherent temporal variations. To address this, we propose SATS, featuring a scale-aware token alignment mechanism that treats patch size as an explicit notion of scale. By incorporating a contrastive-inspired alignment regularizer, SATS aligns representation spaces across scales while preserving distinct modeling capacities. Furthermore, a hybrid masking strategy combining random and contiguous masking is introduced to capture multi-scale temporal structures. Experimental results on LSTF benchmarks demonstrate that SATS achieves a 9.2\% improvement in MSE and an 8.3\% gain in GIFT-Eval MASE compared to competitive baselines. Notably, SATS consistently delivers SOTA performance while achieving a 65.6\% increase in model efficiency over advanced baselines, highlighting its effectiveness and scalability in time series pretraining.
\end{abstract}


\section{Introduction}
The recent emergence of foundation models has significantly advanced various domains such as natural language processing~\cite{GPT-3,Llama-3}, computer vision~\cite{dinov2,CLIP}, and speech understanding~\cite{wav2vec,whisper}. Inspired by their success, growing efforts have been devoted to developing foundation models for time series, aiming to produce general-purpose representations transferable across diverse downstream tasks. An early line of work adapts pretrained language models to time series tasks, leveraging their sequence modeling capabilities in hopes of achieving strong generalization~\cite{Tempo,Time-llm,S2IP-LLM}. However, the modality gap often hinders their performance on temporally structured data, resulting in suboptimal generalization across diverse time series tasks. Moreover, their black-box nature further exacerbates the issue, raising concerns about interpretability and the lack of alignment with intrinsic temporal characteristics~\cite{ARU}. To address these challenges, a second line of work has emerged that trains foundation models from scratch on large-scale, heterogeneous time series datasets~\cite{Time-MoE,Moirai,Chronos}. These models aim to capture universal temporal dynamics in a data-driven and domain-adaptive manner, thereby enhancing robustness to distribution shifts and improving transferability across domains with varying sampling rates, modalities, and sequence lengths (e.g., finance, healthcare, meteorology, IoT).

Despite the promise of the latter direction, it presents unique challenges—
particularly in how to effectively segment and tokenize continuous signals for cross-dataset pretraining. Unlike language, where discrete word units naturally serve as stable tokens~\cite{BPE}, or vision, where uniform patch sizes are viable due to consistent spatial resolution and semantic robustness~\cite{DeiT,ViT}, time series data exhibit irregular sampling and variable sequence lengths, making fixed-size downsampling ineffective. These characteristics necessitate the use of small, adaptive patch sizes to preserve fine-grained temporal patterns.

\begin{figure}[!tbp]
\begin{center}
\centerline{\includegraphics[width=\columnwidth]{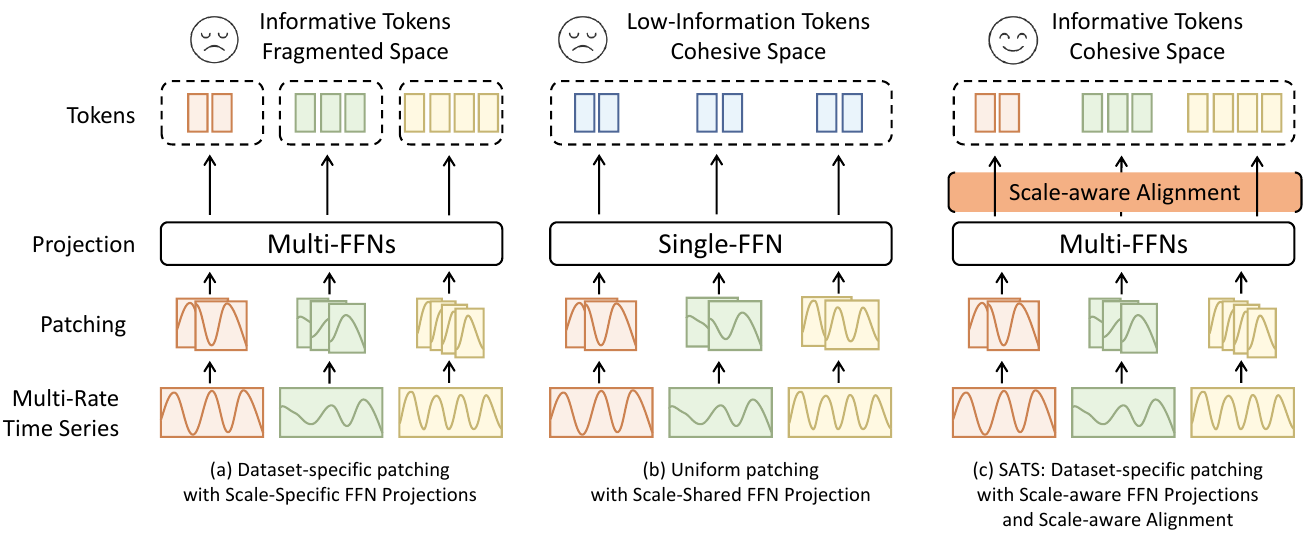}}
\caption{(a) Dataset-specific patch sizes and independent FFNs for varying sampling rates lead to fragmented token spaces. (b) Using a unified patch size and FFN risks information bottlenecks and misaligned local dynamics. (c) SATS adopts dataset-specific patch sizes and enforces scale-aware alignment across FFN-projected spaces, yielding semantically rich and consistent representations.}
\label{intro}
\end{center}
\end{figure}

As shown in Figure~\ref{intro}, recent studies have explored two main strategies for time series tokenization, each with inherent limitations. (1) \textbf{Dataset-specific patching} adopts variable patch sizes tailored to local sampling rates, combined with independent FFNs for token projection~\cite{Elastst,Moirai}. While this design aligns well with the granularity of each dataset, it results in fragmented token spaces that hinder the learning of generalizable temporal patterns and compromise training stability~\cite{Ronen2023VisionTW}. (2) \textbf{Uniform patching} enforces a globally small patch size across datasets to promote representational consistency~\cite{Lightgts,Moirai-moe}. However, this strategy introduces information bottlenecks and often misaligns local dynamics, as it fails to accommodate the diverse temporal structures inherent in different datasets~\cite{TimeMosaic}. Both strategies, therefore, face a trade-off between dataset adaptability and representational generality, limiting their effectiveness in scalable pretraining.

To bridge the gap between fragmented token spaces introduced by adaptive patching and the representational rigidity of fixed segmentation, we propose a scale-aware token alignment mechanism tailored for time series pretraining. By treating the patch size as an explicit notion of scale, our method aligns the representation spaces induced by scale-specific FFNs. This is achieved by minimizing the distance between mean token embeddings across scales to encourage semantic alignment, while simultaneously maximizing the distance between their maximal embeddings to preserve the scale-specific modeling capacity. The resulting token space offers a unified yet expressive foundation for downstream tasks.

Building on this aligned representation space, a remaining challenge lies in the diverse temporal structures inherent to different datasets. Even with aligned embeddings, temporal variations may manifest within individual tokens or span across multiple tokens, depending on the dynamics of the underlying sequence. To capture such variability, we introduce a hybrid masking strategy that enhances multi-scale temporal modeling during masked reconstruction. This strategy combines random masking, which promotes fine-grained inference, with contiguous masking, which facilitates the modeling of long-range dependencies. By jointly optimizing across these complementary patterns, the model learns to recover temporal structures at varying resolutions, improving its robustness and generalization.

Our main contributions are summarized as follows:
\begin{itemize}
    \item We propose \textbf{SATS}, a \textbf{S}cale-\textbf{A}ware foundation model for \textbf{T}ime \textbf{S}eries, which achieves superior generalization across heterogeneous time series datasets.
    \item We introduce a scale-aware alignment mechanism based on scale-specific FFNs, unifying token spaces across patch scales while preserving scale-specific expressiveness.
    \item We design a hybrid masking strategy that enables the model to capture both fine-grained and long-range temporal dependencies across multiple resolutions.
    \item Extensive experiments demonstrate the effectiveness of SATS in both zero-shot and in-distribution forecasting settings, establishing its potential as a strong pretraining paradigm for time series foundation models.
\end{itemize}

\section{Related Work}
\subsection{Time Series Foundation Models}
Traditional forecasting models are typically optimized for specific datasets, horizons, and sampling resolutions, so their scale handling is often encoded through architectural or hyperparameter choices. Multi-scale models such as Pathformer~\cite{Pathformer} and TimeMixer~\cite{TimeMixer} strengthen this line by modeling temporal patterns at multiple resolutions through scale-specific pathways, decomposition, pooling, patching, or dilation. While these methods capture multi-resolution patterns within individual datasets, foundation-model pretraining mixes datasets with different sampling frequencies and patch scales, making cross-frequency representation compatibility necessary for transferable temporal representations. Time series foundation models have therefore gained increasing attention. To meet forecasting-specific demands, decoder-only models such as Timer~\cite{Timer} and Lag-Llama~\cite{Lag-llama} adopt causal architectures, while sparse MoE variants such as Time-MoE~\cite{Time-MoE} and Moirai-MoE~\cite{Moirai-moe} improve scalability. Encoder-decoder models such as LightGTS~\cite{Lightgts} and Chronos~\cite{Chronos} use parallel decoding or discretized objectives. Encoder-only architectures remain less explored for time series foundation models, although recent experimental analyses~\cite{Towards} suggest stronger representational capacity under limited compute, motivating further investigation into their architecture and pretraining strategies~\cite{Moirai,Moment}. Following this encoder-only pretraining direction, SATS addresses the overlooked need to align scale-specific token spaces induced by heterogeneous patch sizes.
\subsection{Contrastive Learning in Pretraining}
Contrastive learning has emerged as a powerful paradigm in large-scale pretraining across various domains. In NLP, methods such as SimCSE~\cite{Simcse} leverage contrastive objectives to learn semantically meaningful sentence embeddings without supervision. In computer vision, CLIP~\cite{CLIP} and ALIGN~\cite{Align} jointly embed images and texts by maximizing the similarity of paired modalities while contrasting unpaired ones, achieving impressive zero-shot performance. While contrastive learning in time series remains relatively underexplored, recent works like TS-TCC~\cite{TS-TCC} and CoST~\cite{CoST} demonstrate its potential in learning transferable representations by aligning augmented views of temporal data. These methods generally seek a balance between aligning semantically related representations and maintaining sufficient dispersion in the embedding space. Such an attraction--repulsion structure not only improves feature discriminability but also helps prevent representations from collapsing into an uninformative configuration. A key advantage of contrastive learning lies in its ability to preserve embedding diversity—by pulling semantically similar instances closer and pushing dissimilar ones apart, it structures the latent space in a discriminative and robust manner. Inspired by contrastive learning’s structured divergence, we adopt an InfoNCE-motivated objective to enhance distinctiveness among multi-scale features—without explicit negative samples—thus inheriting its regularization benefits.
\section{Methodology}
\subsection{Problem Formulation}
Let $\mathcal{S} = \{ (\mathbf{X}^{(i)}, \mathbf{C}^{(i)}) \}_{i=1}^{N}$ denote a dataset of multivariate time series, where $\mathbf{X}^{(i)} \in \mathbb{R}^{d_x \times T_i}$ are target sequences and $\mathbf{C}^{(i)} \in \mathbb{R}^{d_c \times T_i}$ are associated covariates. Given the unmasked observations $\mathbf{X}_{\text{obs}}$ and the corresponding covariates $\mathbf{C}$, the objective is to learn model parameters $\theta$ such that the model $f_\theta$ predicts the distribution parameters $\hat{\boldsymbol{\psi}}$ for the masked subset $\mathbf{X}_{\mathcal{M}}$ of the target sequence.

This leads to the following optimization problem:
\begin{equation}
\begin{aligned}
\min_{\theta}\;&
\mathbb{E}_{(\mathbf{X}, \mathbf{C}) \sim p(\mathcal{S})}
\mathbb{E}_{\mathcal{M} \sim p(\mathcal{T}\mid\mathcal{S})}
\left[
\mathcal{L}_{\text{nll}}
\left(\mathbf{X}_{\mathcal{M}}, \hat{\boldsymbol{\psi}}\right)
\right] \\
\text{s.t.}\;&
\hat{\boldsymbol{\psi}} = f_\theta(\mathbf{X}_{\text{obs}}, \mathbf{C})
\end{aligned}
\end{equation}
Here, $\mathcal{L}_{\text{nll}}$ denotes the \textit{negative log-likelihood loss}:
\begin{equation}\label{eq:nll}
\mathcal{L}_{\text{nll}}(\mathbf{X}_{\mathcal{M}}, \hat{\boldsymbol{\psi}}) = -\log p(\mathbf{X}_{\mathcal{M}} \mid \hat{\boldsymbol{\psi}})
\end{equation}
where $p(\mathcal{S})$ is the data-generating distribution over time series instances $(\mathbf{X}, \mathbf{C})$, and $p(\mathcal{T} \mid \mathcal{S})$ defines the task sampling distribution that governs the selection of masked positions $\mathcal{M} \subset \{1, \dots, T\}$ for prediction. Classical forecasting corresponds to the special case where the masked region $\mathcal{M}$ is located at the end of the sequence.

\begin{figure*}[!tbp]
\begin{center}
\centerline{\includegraphics[width=\textwidth]{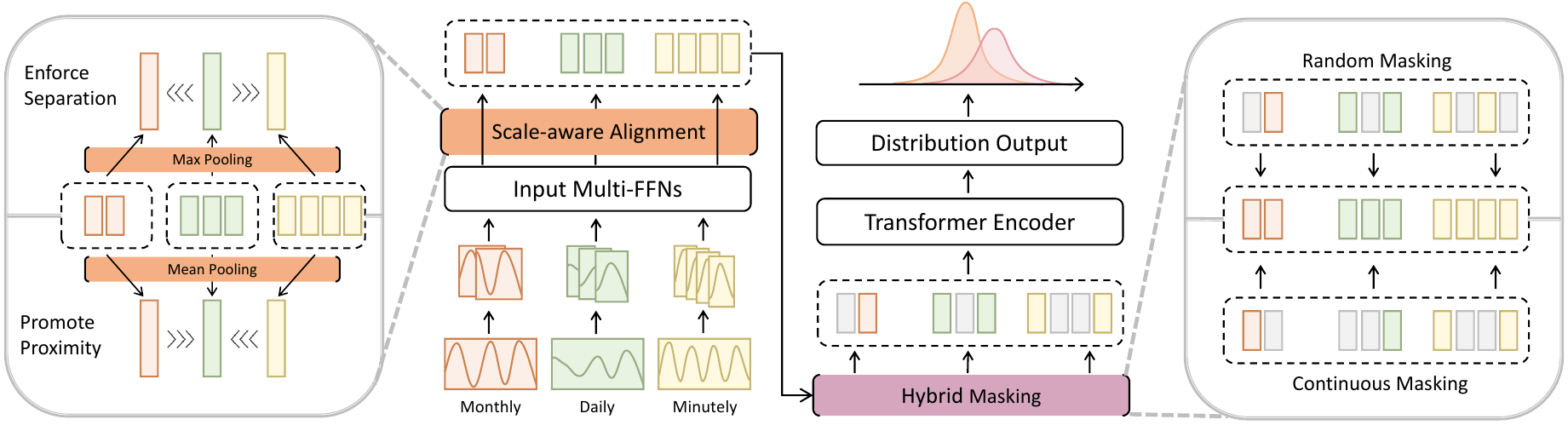}}
\caption{Overview of the SATS framework. Tokens from multiple patch sizes are projected via separate FFNs. SATS employs \textbf{Scale-aware Alignment} mechanism to promote proximity of mean-pooled representations within each scale, while enforcing separation of max-pooled representations across scales—balancing consistency and scale-specific expressiveness. \textbf{Hybrid masking strategy}, integrating Random Masking and Continuous Masking, is further applied to capture both fine-grained and long-range temporal dependencies.}
\label{frame}
\end{center}
\end{figure*}
\subsection{Model Architecture}

As shown in Figure~\ref{frame}, SATS adopts a non-overlapping patch-based, encoder-only Transformer~\cite{PatchTST}. The multivariate time series is first flattened and, following Moirai~\cite{Moirai}, mapped into patches of varying sizes based on the dataset. To improve efficiency, we adopt packing as a default setting~\cite{packing,Llama-3}, enabling tokens with different patch sizes from multiple datasets to be packed into a single sequence. This multi-scale design introduces inconsistencies in the token space; while packing is not the direct cause, it is an indispensable component of modern scalable training, making it both practical and necessary to develop solutions within this paradigm. 

To mitigate such inconsistencies while embracing the packing paradigm, SATS employs a scale-aware alignment mechanism: it pulls closer the mean-pooled representations within the same scale, while pushing apart the max-pooled ones across scales, ensuring consistency while preserving scale-specific expressiveness. Based on this aligned space, a hybrid masking strategy combining random and contiguous patterns is applied to capture both fine-grained and long-range dependencies.

Although not shown, the encoder incorporates key techniques from foundational model pretraining—such as RoPE~\cite{RoPE}, SwiGLU~\cite{SwiGLU}, and RMSNorm~\cite{RMSNorm}—as well as inductive biases specific to time-series pretraining, including Any-Variate Bias, Mixture Distribution Output~\cite{Moirai} and RevIN~\cite{RevIN} for modeling inter-variable dependencies and normalization under distribution shifts.

\subsubsection{Scale-aware Alignment}\label{Sec-sa}  
To enhance the effectiveness of universal temporal modeling, especially when dealing with subsequences of varying scales, it is crucial to design a robust embedding space alignment strategy. Given token sequences \( \mathcal{I} \in \mathbb{R}^{L \times D} \), where \(L\) represents the maximum input length during training and \(D\) is the hidden layer dimension of the encoder, the challenge arises from the coexistence of tokens originating from \(n \leq N\) different patch sizes, where \(N\) denotes the total number of distinct patch sizes. A direct approach could be to minimize the feature space distance, such as cosine similarity, between subsequences, encouraging their proximity. However, this approach faces several challenges: first, the varying lengths of subsequences make it difficult to quantify alignment; second, different samples within the same batch may contain different numbers of subsequences, complicating the application of proximity constraints both within and across samples. Furthermore, a structured information constraint is necessary to discourage scale-wise anchor collapse and maintain diverse temporal representations.

In response to these challenges, we propose the Scale-aware Alignment (SA) method, which integrates two key components. First, we introduce a pooling mechanism to address the issues of variable subsequence lengths and differing numbers of subsequences across samples. Specifically, we pool tokens based on their patch sizes to generate the scale-wise embedding representation $Y \in \mathbb{R}^{N \times D}$.  In cases where a patch size is absent in a given sample, the corresponding embedding position $Y_i$ is set to zero ($i \leq N$) to maintain fixed tensor dimensions for batching. Crucially, a binary validity mask is applied during the computation of the alignment loss to explicitly exclude these zeroed positions from the similarity summation, thereby preventing gradient propagation from missing patch sizes. Second, inspired by the principles of contrastive learning, we design a structured information constraint: the mean embeddings from different patch sizes are pulled closer to establish neighboring centers in the token space, while the maximal embeddings are repelled to encode scale-specific information, ensuring richer and more diverse token semantics. More theoretical analysis is provided in Section~\ref{Sec-theory-minmax}. To operationalize this constraint, we adopt the InfoNCE framework, as detailed in Equation~\ref{eq:close} and Equation~\ref{eq:far}, where $ \cos(\cdot) $ denotes the cosine similarity function and $ \tau $ is the temperature parameter.
\begin{equation}\label{eq:close}
\begin{aligned}
\mathcal{L}_\text{close}
= -\mathbb{E}\!\left[
\log
\frac{\sum_{j \neq i} \exp \left( \cos(Y_i \cdot Y_j) /\tau \right)}
{\sum_{j=1}^N \exp \left( \cos(Y_i \cdot Y_j) /\tau \right)}
\right]
\end{aligned}
\end{equation}
\begin{equation}\label{eq:far}
\begin{aligned}
\mathcal{L}_\text{far}
= -\mathbb{E}\!\left[
1 - \log \left(
\sum_{j=1}^N \exp \left( \cos(Y_i \cdot Y_j)/\tau \right)
\right)
\right]
\end{aligned}
\end{equation}

In practice, $ Y_i \in Y^{\text{mean}} $ is sequentially substituted into Equation~\ref{eq:close}, while $ Y_i \in Y^{\text{max}} $ is substituted into Equation~\ref{eq:far}. Although both equations follow the InfoNCE form, they do not involve true negative samples. We therefore combine these two losses to form the final scale-aware alignment constraint in Equation~\ref{eq:sa}. This design ensures that the loss function conforms to the geometric structure of contrastive learning, as detailed in Section~\ref{Sec-theory-geo}. Consequently, it provides structured regularization that aligns feature representations across different patch sizes, enhances cross-scale consistency, and discourages scale-wise anchor collapse. The hyperparameter $\beta$ controls the relative weight of the maximal embedding pull-away term, balancing the overall objective.
\begin{equation}\label{eq:sa}
\mathcal{L}_\text{sa}=\mathcal{L}_\text{close}+\beta \mathcal{L}_\text{far}
\end{equation}

\subsubsection{Hybrid Masking Strategy}
On top of the aligned token space, the intrinsic heterogeneity and complexity of temporal dynamics across datasets continue to challenge effective representation learning. Although alignment mitigates certain variations, temporal dependencies inherently span multiple scales: some manifest as fine-grained, localized fluctuations within individual tokens, while others emerge as extended, structured patterns across contiguous token segments. To comprehensively capture these diverse temporal scales and improve the robustness of learned representations, we therefore propose a Hybrid Masking Strategy (HM) that synergistically combines Random Masking (RM) with Continuous Masking (CM), implemented as contiguous-span masking during pretraining.

Concretely, given each token subsequence \(\mathcal{I}_j \in \mathbb{R}^{L_j \times D}\) extracted from the full sequence \(\mathcal{I}\), where $L_j$ denotes the length of the $j$-th subsequence, a masking ratio \(r \in [0.15, 0.5]\) is applied. For each subsequence, a predefined probability \(p \in [0,1]\) determines whether RM or CM is used.
With probability \(p\), RM uniformly selects \(m_j\) token positions, where \(m_j = \lceil r \cdot L_j \rceil\) and \(\sum_{i=0}^{L_j-1} \mathcal{M}_r^{(j)}(i)=m_j\), producing a binary mask \(\mathcal{M}_r^{(j)}\):
\begin{equation}
\mathcal{M}_r^{(j)}(i) = 
\begin{cases}
1, & \text{if token } i \text{ is randomly selected}  \\
0, & \text{otherwise}
\end{cases}.
\end{equation}
Note that $m_j \le L_j$ is inherently satisfied given $r \le 0.5$. Alternatively, with probability \(1 - p\), CM samples a start index \(s_j \in \{0, \ldots, L_j - m_j\}\), masking a continuous block of tokens:
\begin{equation}
\mathcal{M}_c^{(j)}(i) = 
\begin{cases}
1, & s_j \leq i < s_j + m_j \\
0, & \text{otherwise}.
\end{cases}
\end{equation}
The final mask \(\mathcal{M}^{(j)}\) applied to each subsequence \(\mathcal{I}_j\) is sampled as
\begin{equation}\label{eq:full_mask}
\mathcal{M}^{(j)} = 
\begin{cases}
\mathcal{M}_r^{(j)}, & \text{with probability } p \\
\mathcal{M}_c^{(j)}, & \text{with probability } 1 - p.
\end{cases}
\end{equation}

By guiding the model to recover masked tokens across both RM and CM patterns, HM balances fine-grained local inference and long-range dependency learning. Consequently, it enhances the robustness and generalizability of learned representations for diverse temporal modeling tasks.

\subsection{Model Training}
\subsubsection{Unified Learning Objective}
The proposed alignment and masking strategies introduce no additional learnable parameters, enabling their integration into a unified learning objective without parameter overhead. In practice, the mask $\mathcal{M}$ obtained from Equation~\ref{eq:full_mask} is applied to Equation~\ref{eq:nll} to compute the primary training loss. Simultaneously, Equation~\ref{eq:sa} is employed as an auxiliary training loss to enforce SA. We combine them into the total loss function as follows:
\begin{equation}
    \mathcal{L} = \mathcal{L}_{\text{nll}} + \alpha \mathcal{L}_{\text{sa}}
\end{equation}
where $\alpha$ is a weighting coefficient balancing the two objectives.

\subsubsection{SATS Setup}
We pretrain the SATS models on the LOTSA dataset~\cite{Moirai} in two configurations—small and base—with detailed model specifications provided in Table~\ref{tab:size}. The small model is trained for 100{,}000 steps with a batch size of 256, while the base model is trained for 200{,}000 steps with a batch size of 128. All experiments use the following fixed hyperparameters unless otherwise specified:

\begin{itemize}
\item Optimizer: AdamW with learning rate $1 \times 10^{-3}$, weight decay $1 \times 10^{-1}$, $\beta_1 = 0.9$, $\beta_2 = 0.98$.
\item SA: Temperature $\tau_{\text{mean}} = 0.1$ (Eq.~\ref{eq:close}), $\tau_{\text{max}} = 0.2$ (Eq.~\ref{eq:far}).
\item HM: Masking probability $p = 0.5$ for balanced RM and CM.
\item Loss Weights: Primary objective weight $\alpha = 0.1$, auxiliary objective weight $\beta = 0.3$.
\end{itemize}

Due to limited computational resources and empirical evidence suggesting that large-scale language model pretraining is relatively robust to hyperparameter choices within reasonable ranges — as performance is primarily governed by scale rather than fine-tuned hyperparameters \cite{RoBERTa, scaleLawsLLM} — no further hyperparameter tuning was performed beyond the values listed above.
\begin{table}[!tbp]
\centering
\caption{Key parameter details of SATS model sizes.}
\label{tab:size}
\begin{tabular}{cccccc}
\hline
 & Layers & $\text{d}_\text{model}$ & $\text{d}_\text{ff}$ & Heads & Params \\ \hline
$\text{SATS}_\text{S}$ & 6 & 384 & 1536 & 6 & 14M \\
$\text{SATS}_\text{B}$  & 9 & 768 & 3072 & 12 & 70M \\ \hline
\end{tabular}
\end{table} 

\section{Theoretical Analysis} \label{Sec-theory}
\subsection{Mean Consistency and Max Discriminability} \label{Sec-theory-minmax}
We provide a concise justification for the two pooling choices in SA: mean-pooled representations are pulled together to encourage cross-scale semantic consistency, whereas max-pooled representations are pushed apart to preserve scale-specific information.

Consider a time series patch decomposed as
\[
    x[n] = \ell[n] + h[n],
\]
where $\ell[n]$ is a low-frequency trend satisfying the Lipschitz condition $|\ell[n]-\ell[m]|\le K|n-m|$~\cite{giraud2015aggregation,yakowitz1999strongly}, and $h[n]$ is a zero-mean high-frequency component with bounded variance and short-range dependence. For a patch of length $P$, define
\[
    \mu_P = \frac{1}{P}\sum_{n=1}^{P} x[n],
    \qquad
    M_P = \max_{1\le n\le P} x[n].
\]

\subsubsection{Mean consistency}
The mean statistic suppresses high-frequency fluctuations. Under short-range dependence,
\[
    \mathrm{Var}(\mu_P)
    =
    \frac{1}{P^2}\sum_{n,m=1}^{P}\mathrm{Cov}(h[n],h[m])
    \le
    \frac{C_{\mathrm{cov}}\sigma_{\max}^2}{P},
\]
which shows that the variance of the high-frequency component decays as $O(1/P)$. Moreover, since the trend component is Lipschitz smooth, the expectation gap between two patch lengths is controlled by their scale difference:
\[
    \big|\mathbb{E}[\mu_{P_1}] - \mathbb{E}[\mu_{P_2}]\big|
    \le \frac{K}{2}|P_1-P_2|.
\]
Thus, mean pooling yields stable scale-wise prototypes that mainly retain shared low-frequency semantics.

\subsubsection{Max discriminability}
In contrast, the max statistic remains sensitive to local high-frequency responses. Under mild tail conditions for the high-frequency component, classical extreme value theory~\cite{leadbetter2012extremes} gives
\[
    \mathbb{E}\left[\max_{n\le P} h[n]\right]
    \asymp
    \sigma_{\max}\sqrt{2\log P},
\]
up to lower-order terms. Therefore, max-pooled anchors emphasize scale-dependent peaks whose magnitude varies with the patch length. This makes them suitable for preserving scale-specific information that would be weakened if all scale representations were only pulled together.

\subsubsection{Summary}
These complementary properties motivate the design of SA. Pulling mean-pooled embeddings together promotes a coherent cross-scale semantic space, while pushing max-pooled embeddings apart preserves scale-sensitive anchor diversity and mitigates scale-wise representation collapse.

\subsection{Geometric Correspondence with Contrastive Learning}\label{Sec-theory-geo}

We further examine the intrinsic connection between SA and contrastive representation learning. Although our objective does not rely on conventional instance-level negative sampling, it preserves the core contrastive geometry: aligning related representations while dispersing informative anchors. This perspective explains why mean-attraction and max-repulsion are consistent with contrastive learning principles and can promote cross-scale consistency while reducing collapse risk.

\subsubsection{Structured contrastive geometry}
We make this geometric prior explicit through a normalized L2 analytic surrogate:
\[
    \mathcal{L}_{\text{sa}}^{\text{analytic}}
    =
    \underbrace{\sum_{i=1}^N \|h_i - \bar h\|^2}_{\mathcal{L}_{\text{align}}}
    -
    \beta
    \underbrace{\sum_{i \neq t} \|m_i - m_t\|^2}_{\mathcal{L}_{\text{sep}}},
\]
where $\{m_i\}_{i=1}^N \subset \mathbb{S}^{D-1}$ are normalized scale-wise max anchors, $h_i$ are scale-wise mean prototypes, and $\bar h$ is their centroid. For normalized embeddings, $\|x-y\|^2 = 2 - 2x^\top y$; hence the first term increases similarity between mean prototypes and a shared semantic center, while the second decreases similarity among max anchors. The surrogate does not replace the InfoNCE-style training loss; it exposes the same normalized attraction--repulsion geometry. Equations~\ref{eq:close}--\ref{eq:far} implement this geometry in smooth softmax form: mean-pooled prototypes are aligned, and scale-dominant max anchors are dispersed. This correspondence yields a structured alignment--uniformity variant~\cite{wang2022understandingcontrastiverepresentationlearning}, where uniformity acts only on the compact anchor set $\{m_i\}$ rather than all tokens.

\subsubsection{Mitigation of Representation Collapse} 
The separation term provides an anti-collapse bias for scale-wise max anchors. Since these anchors capture scale-sensitive high-variation responses, angular separation preserves diversity in the scale-anchor subspace. This connects to dimensional collapse, commonly characterized by degeneration of the embedding covariance spectrum~\cite{jing2022understandingdimensionalcollapsecontrastive}.

Formally, let $\{m_i\}_{i=1}^N \subset \mathbb{S}^{D-1}$ denote the normalized max-pooled scale anchors, with $\bar m=N^{-1}\sum_i m_i$ and
\[
    C_m=\frac{1}{N}\sum_{i=1}^N(m_i-\bar m)(m_i-\bar m)^\top .
\]
Then
\[
    \mathrm{tr}(C_m)
    =1-\|\bar m\|^2
    =\frac{1}{2N^2}\sum_{i,j=1}^N\|m_i-m_j\|^2 .
\]
Thus, increasing pairwise separation directly increases total anchor variance. Consequently, $\mathcal{L}_{\text{sep}}$ complements $\mathcal{L}_{\text{align}}$ by discouraging scale-wise point collapse and preserving non-zero variance in the anchor subspace.

\subsubsection{Summary} 
Overall, $\mathcal{L}_{\text{sa}}$ is a lightweight anchor-level contrastive regularizer. It aligns mean-pooled prototypes for cross-scale consistency, disperses max-pooled anchors for scale-specific diversity, and reduces scale-wise collapse risk without large-batch negative sampling.

\section{Experiments}

\subsection{Benchmark Description}
We evaluate SATS on three complementary benchmark collections: LSTF~\cite{Informer}, GIFT-Eval~\cite{GIFT-Eval}, and Monash~\cite{Monash}. Together, they cover long-horizon forecasting, zero-shot cross-dataset generalization, and in-distribution evaluation across diverse domains, frequencies, and horizons.

\begin{itemize}
\item \textbf{LSTF}: A collection of widely used multivariate datasets from domains such as electricity, traffic, and weather, used here to evaluate long-term forecasting.

\item \textbf{GIFT-Eval}: A general forecasting benchmark for evaluating transfer to unseen heterogeneous datasets. It spans diverse domains, frequencies, variable types, and prediction horizons, and reports both point and probabilistic forecasting quality through metrics such as MASE and CRPS.

\item \textbf{Monash}: A large-scale collection of univariate real-world datasets, used here for in-distribution evaluation under heterogeneous data distributions.
\end{itemize}

\begin{table*}[!tbp]
\caption{Full results of zero-shot forecasting across all evaluated models. Lower values of MSE and MAE indicate superior performance. As TimesFM incorporates Weather data during pretraining, it is excluded from evaluation on this dataset (denoted by “–”). \boldres{Red} highlights the best result, while \secondres{Blue} marks the second best. 
}
\label{tab:zero}
\resizebox{\textwidth}{!}{%
\begin{tabular}{cc|cccccccccccccccc}
\hline
\multicolumn{2}{c|}{Models} & \multicolumn{2}{c}{$\text{SATS}_\text{S}$} & \multicolumn{2}{c}{$\text{SATS}_\text{B}$} & \multicolumn{2}{c}{Timer-XL} & \multicolumn{2}{c}{$\text{Time-MoE}_\text{B}$} & \multicolumn{2}{c}{$\text{Moirai}_\text{B}$} & \multicolumn{2}{c}{$\text{Chronos}_\text{L}$} & \multicolumn{2}{c}{Moment} & \multicolumn{2}{c}{TimesFM} \\ \hline
\multicolumn{2}{c|}{Metrics} & MSE & MAE & MSE & MAE & MSE & MAE & MSE & MAE & MSE & MAE & MSE & MAE & MSE & MAE & MSE & MAE \\ \hline
\multicolumn{1}{c|}{\multirow{5}{*}{\rotatebox[origin=c]{90}{ETTh1}}} & 96 & 0.375 & 0.393 & {\secondres{0.360}} & \secondres{0.387} & 0.369 & 0.391 & \boldres{0.357} & {\boldres{0.381}} & 0.383 & 0.402 & 0.441 & 0.390 & 0.688 & 0.557 & 0.414 & 0.404 \\
\multicolumn{1}{c|}{} & 192 & 0.412 & 0.415 & {\secondres{0.395}} & {\secondres{0.409}} & 0.405 & 0.413 & \boldres{0.384} & \boldres{0.404} & 0.425 & 0.429 & 0.502 & 0.424 & 0.688 & 0.560 & 0.465 & 0.434 \\
\multicolumn{1}{c|}{} & 336 & 0.423 & 0.425 & {\secondres{0.413}} & \boldres{0.422} & 0.418 & {\secondres{0.423}} & \boldres{0.411} & 0.434 & 0.456 & 0.450 & 0.576 & 0.467 & 0.675 & 0.563 & 0.503 & 0.456 \\
\multicolumn{1}{c|}{} & 720 & {\secondres{0.418}} & {\secondres{0.441}} & \boldres{0.413} & \boldres{0.438} & 0.423 & {\secondres{0.441}} & 0.449 & 0.477 & 0.470 & 0.473 & 0.835 & 0.583 & 0.683 & 0.585 & 0.511 & 0.481 \\
\multicolumn{1}{c|}{} & AVG & 0.407 & 0.418 & \boldres{0.395} & \boldres{0.414} & 0.404 & {\secondres{0.417}} & {\secondres{0.400}} & 0.424 & 0.433 & 0.438 & 0.589 & 0.466 & 0.684 & 0.566 & 0.473 & 0.444 \\ \hline
\multicolumn{1}{c|}{\multirow{5}{*}{\rotatebox[origin=c]{90}{ETTh2}}} & 96 & {\secondres{0.283}} & \secondres{0.328} & \boldres{0.273} & 0.331 & {\secondres{0.283}} & 0.342 & 0.305 & 0.359 & 0.277 & {\boldres{0.327}} & 0.320 & 0.345 & 0.342 & 0.396 & 0.315 & 0.349 \\
\multicolumn{1}{c|}{} & 192 & 0.343 & \boldres{0.369} & \boldres{0.330} & {\secondres{0.372}} & {\secondres{0.340}} & 0.379 & 0.351 & 0.386 & {\secondres{0.340}} & 0.374 & 0.406 & 0.399 & 0.354 & 0.402 & 0.388 & 0.395 \\
\multicolumn{1}{c|}{} & 336 & 0.365 & \boldres{0.391} & \boldres{0.353} & {\secondres{0.396}} & 0.366 & 0.400 & 0.391 & 0.418 & 0.371 & 0.401 & 0.492 & 0.453 & {\secondres{0.356}} & 0.407 & 0.422 & 0.427 \\
\multicolumn{1}{c|}{} & 720 & {\secondres{0.404}} & {\secondres{0.424}} & \boldres{0.380} & \boldres{0.409} & 0.397 & 0.431 & 0.419 & 0.454 & 0.394 & 0.426 & 0.603 & 0.511 & 0.395 & 0.434 & 0.443 & 0.454 \\
\multicolumn{1}{c|}{} & AVG & 0.349 & {\secondres{0.378}} & \boldres{0.334} & \boldres{0.377} & {\secondres{0.347}} & 0.388 & 0.367 & 0.404 & 0.345 & 0.382 & 0.455 & 0.427 & 0.362 & 0.410 & 0.392 & 0.406 \\ \hline
\multicolumn{1}{c|}{\multirow{5}{*}{\rotatebox[origin=c]{90}{ETTm1}}} & 96 & 0.325 & \secondres{0.353} & \secondres{0.323} & \boldres{0.345} & \boldres{0.317} & 0.356 & 0.338 & 0.368 & 0.396 & 0.382 & 0.457 & 0.403 & 0.654 & 0.527 & 0.361 & 0.370 \\
\multicolumn{1}{c|}{} & 192 & \boldres{0.352} & {\secondres{0.372}} & \boldres{0.352} & \boldres{0.364} & 0.358 & 0.381 & {\secondres{0.353}} & 0.388 & 0.425 & 0.402 & 0.530 & 0.450 & 0.662 & 0.532 & 0.414 & 0.405 \\
\multicolumn{1}{c|}{} & 336 & {\secondres{0.372}} & {\secondres{0.387}} & \boldres{0.371} & \boldres{0.379} & 0.386 & 0.401 & 0.381 & 0.413 & 0.452 & 0.415 & 0.577 & 0.481 & 0.672 & 0.537 & 0.445 & 0.429 \\
\multicolumn{1}{c|}{} & 720 & {\secondres{0.405}} & {\secondres{0.410}} & \boldres{0.401} & \boldres{0.403} & 0.430 & 0.431 & 0.504 & 0.493 & 0.477 & 0.431 & 0.660 & 0.526 & 0.692 & 0.551 & 0.512 & 0.471 \\
\multicolumn{1}{c|}{} & AVG & {\secondres{0.364}} & {\secondres{0.380}} & \boldres{0.362} & \boldres{0.373} & 0.373 & 0.392 & 0.394 & 0.416 & 0.437 & 0.407 & 0.556 & 0.465 & 0.670 & 0.537 & 0.433 & 0.419 \\ \hline
\multicolumn{1}{c|}{\multirow{5}{*}{\rotatebox[origin=c]{90}{ETTm2}}} & 96 & {\secondres{0.172}} & {\secondres{0.255}} & \boldres{0.167} & \boldres{0.251} & 0.189 & 0.277 & 0.201 & 0.291 & 0.195 & 0.269 & 0.197 & 0.271 & 0.260 & 0.335 & 0.202 & 0.270 \\
\multicolumn{1}{c|}{} & 192 & {\secondres{0.226}} & {\secondres{0.292}} & \boldres{0.222} & \boldres{0.290} & 0.241 & 0.315 & 0.258 & 0.334 & 0.247 & 0.303 & 0.254 & 0.314 & 0.289 & 0.350 & 0.289 & 0.321 \\
\multicolumn{1}{c|}{} & 336 & {\secondres{0.279}} & {\secondres{0.327}} & \boldres{0.269} & \boldres{0.323} & 0.286 & 0.348 & 0.324 & 0.373 & 0.291 & 0.333 & 0.313 & 0.353 & 0.324 & 0.369 & 0.360 & 0.366 \\
\multicolumn{1}{c|}{} & 720 & 0.369 & 0.385 & \boldres{0.343} & \boldres{0.374} & 0.375 & 0.402 & 0.488 & 0.464 & {\secondres{0.355}} & {\secondres{0.377}} & 0.416 & 0.415 & 0.394 & 0.409 & 0.462 & 0.430 \\
\multicolumn{1}{c|}{} & AVG & {\secondres{0.262}} & {\secondres{0.315}} & \boldres{0.250} & \boldres{0.309} & 0.273 & 0.336 & 0.318 & 0.366 & 0.272 & 0.321 & 0.295 & 0.338 & 0.317 & 0.366 & 0.328 & 0.347 \\ \hline
\multicolumn{1}{c|}{\multirow{5}{*}{\rotatebox[origin=c]{90}{Weather}}} & 96 & 0.180 & 0.236 & {\secondres{0.162}} & 0.217 & 0.171 & 0.225 & \boldres{0.160} & \secondres{0.214} & 0.176 & \boldres{0.210} & 0.194 & 0.235 & 0.243 & 0.255 & - & - \\
\multicolumn{1}{c|}{} & 192 & 0.226 & 0.280 & \boldres{0.210} & 0.265 & 0.221 & 0.271 & \boldres{0.210} & {\secondres{0.260}} & {\secondres{0.218}} & \boldres{0.251} & 0.249 & 0.285 & 0.278 & 0.329 & - & - \\
\multicolumn{1}{c|}{} & 336 & 0.274 & 0.316 & \boldres{0.258} & {\secondres{0.302}} & 0.274 & 0.311 & 0.274 & 0.309 & {\secondres{0.267}} & \boldres{0.288} & 0.302 & 0.327 & 0.306 & 0.346 & - & - \\
\multicolumn{1}{c|}{} & 720 & 0.341 & 0.363 & \boldres{0.325} & {\secondres{0.349}} & 0.356 & 0.370 & 0.418 & 0.405 & {\secondres{0.338}} & \boldres{0.338} & 0.372 & 0.378 & 0.350 & 0.374 & - & - \\
\multicolumn{1}{c|}{} & AVG & 0.255 & 0.299 & \boldres{0.239} & {\secondres{0.283}} & 0.256 & 0.294 & 0.266 & 0.297 & {\secondres{0.250}} & \boldres{0.271} & 0.279 & 0.306 & 0.294 & 0.326 & - & - \\ \hline
\multicolumn{2}{c|}{Average} & {\secondres{0.327}} & {\secondres{0.358}} & \boldres{0.316} & \boldres{0.351} & 0.330 & 0.365 & 0.349 & 0.381 & 0.348 & 0.364 & 0.435 & 0.401 & 0.465 & 0.441 & - & - \\ \hline
\multicolumn{2}{c|}{$1^{st}$ Count} & \multicolumn{2}{c}{3} & \multicolumn{2}{c}{\boldres{35}} & \multicolumn{2}{c}{1} & \multicolumn{2}{c}{\secondres{7}} & \multicolumn{2}{c}{6} & \multicolumn{2}{c}{0} & \multicolumn{2}{c}{0} & \multicolumn{2}{c}{0} \\ \hline
\end{tabular}
}
\end{table*}

\subsection{Benchmarking Setup}
\textbf{Baselines.} We conduct extensive comparisons with widely adopted time series foundation models, including Timer-XL~\cite{Timer-XL}, Time-MoE~\cite{Time-MoE}, Moirai~\cite{Moirai}, Chronos~\cite{Chronos}, Moment~\cite{Moment}, TimesFM~\cite{TimesFM}, and TTM-R2~\cite{TTMs}. To broaden the comparative scope, we additionally incorporate two methods adapted from foundation models of other modalities: VisionTS~\cite{VisionTS} and LLMTime~\cite{LLMTime}. Furthermore, following the recommendations of Bergmeir~\cite{NIPSworkshop}, we extend our in-distribution evaluation to encompass traditional baselines, including Naive, ETS~\cite{ETS}, and DeepAR~\cite{DeepAR}.

\textbf{Evaluation Setup.} Baselines follow their original configurations. For SATS, we follow Moirai, search context lengths in \{1000, 2000, 3000, 4000, 5000\}, and use Woo et al.'s frequency-specific patch candidates~\cite{Moirai}:
\begin{itemize}
    \item Yearly/Quarterly: 8
    \item Monthly: 8, 16, 32
    \item Weekly/Daily: 16, 32
    \item Hourly: 32, 64
    \item Minute-level: 32, 64, 128
    \item Second-level: 64, 128
\end{itemize}
Empirically, we choose the largest feasible patch size with a lookback of at least 3000. Metrics use 100 predictive samples and report results based on the mean prediction. As pretrained baselines often impose model-specific input lengths, such as Time-MoE's four-horizon input~\cite{Time-MoE} and Timer-XL's dataset-dependent lengths~\cite{Timer-XL}, we follow each model's recommended setting for fairness.

\begin{figure*}[!t]
\begin{center}
\centerline{\scalebox{1}[0.88]{\includegraphics[width=\textwidth]{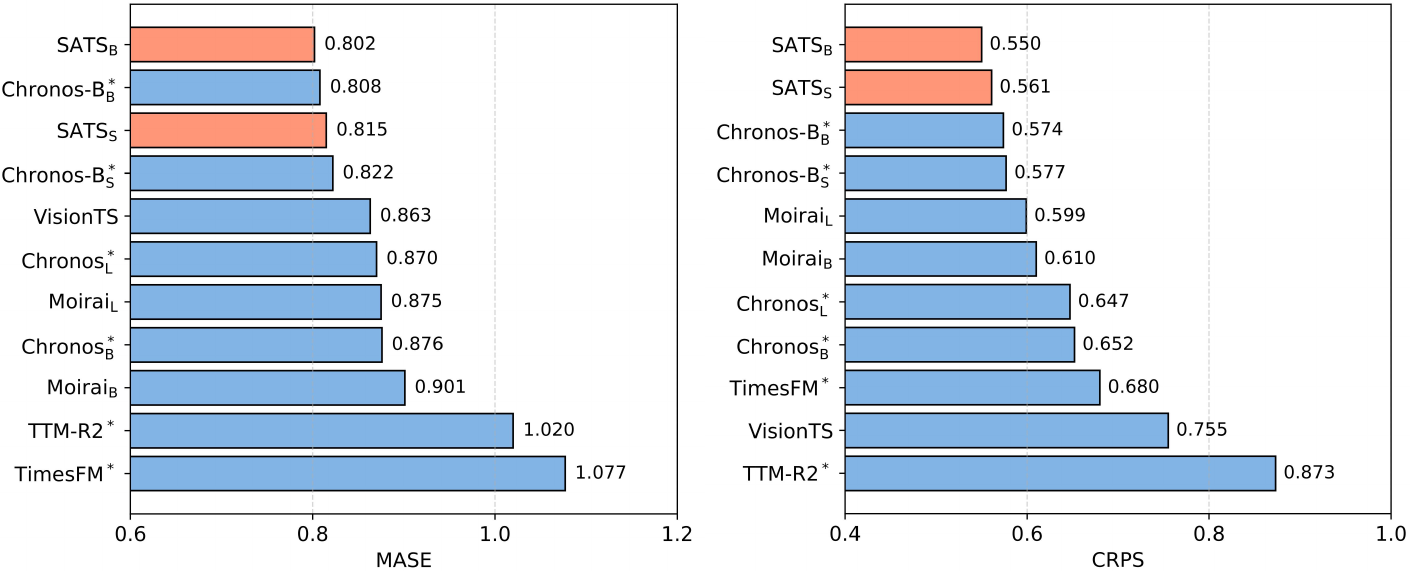}}}
\caption{Zero-shot forecasting performance evaluated on 23 datasets from GIFT-Eval benchmark~\cite{GIFT-Eval}. Lower values of MASE and CRPS indicate superior performance. Methods trained with access to these evaluation datasets during pretraining are denoted with asterisks (*). Results were aggregated in accordance with the standard method of GIFT-Eval.
}
\label{fig:GE}
\end{center}
\end{figure*}

\subsection{Zero-shot Forecasting}
We evaluate zero-shot forecasting on LSTF and GIFT-Eval. LSTF tests long-horizon forecasting on datasets excluded from pretraining, while GIFT-Eval tests cross-dataset generalization under a leakage-controlled protocol.
\subsubsection{LSTF Benchmark}

We evaluate five LSTF datasets excluded from LOTSA, using horizons $\{96, 192, 336, 720\}$ and reporting MSE and MAE. For baselines with multiple variants, we exclude models above 1B parameters and report the variant with the best average performance. Electricity, Traffic, and PEMS are excluded because they appear in the pretraining corpora of most models~\cite{Timer-XL,Time-MoE,Moirai}.

\textbf{Result.}
The detailed zero-shot results are presented in Table~\ref{tab:zero}, where $\text{SATS}_\text{B}$ consistently achieves state-of-the-art performance. Compared to $\text{Moirai}_\text{B}$ (encoder-only), the strongest multi-scale FFNs-embedded baseline, $\text{SATS}_\text{B}$ achieves a \textbf{9.2\%} improvement in MSE. It also outperforms Timer-XL (decoder-only) and $\text{Chronos}_\text{L}$ (encoder-decoder) with MSE improvements of \textbf{4.2\%} and \textbf{27.4\%}, respectively. Notably, $\text{SATS}_\text{B}$ contains only 70M parameters, which is substantially fewer than those of the compared baselines. Moreover, even the lightweight $\text{SATS}_\text{S}$ with 14M parameters surpasses all other baselines in overall average performance, highlighting its efficiency. In addition, SATS exhibits a clear performance gain as model size increases, revealing strong scalability. This trend contrasts with models such as Time-MoE~\cite{Time-MoE} and Moirai~\cite{Moirai}, whose performance plateaus or even degrades with larger model configurations.

\subsubsection{GIFT-Eval Benchmark}
We follow the standard GIFT-Eval protocol on 23 datasets spanning economics, energy, healthcare, nature, sales, transportation, and cloud operations. To enforce a strict zero-shot setting, we remove all GIFT-Eval overlaps from LOTSA and retrain SATS. Performance is aggregated by the official protocol, covering both point and probabilistic forecasting metrics.

\textbf{Result.}
The detailed zero-shot results are shown in Figure~\ref{fig:GE}. Overall, $\text{SATS}_\text{B}$ delivers consistently state-of-the-art performance across metrics. Against the strongest encoder-only baseline, $\text{Moirai}_\text{L}$, it yields substantial gains—\textbf{8.3\%} in MASE and \textbf{8.2\%} in CRPS. Notably, even when set against the runner-up, $\text{Chronos-B}_\text{B}$, SATS maintains a clear lead in CRPS under a strictly leakage-free evaluation protocol while requiring only \textbf{35\%} of its parameters, underscoring its ability to learn highly transferable temporal representations during pretraining. Beyond encoder-based models, SATS also surpasses VisionTS—which converts sequences into images for training—with consistently more stable improvements on both MASE and CRPS. This suggests that SATS captures the underlying data distribution more faithfully, rather than merely improving point forecasts as VisionTS tends to do. Together, these findings highlight SATS’s robust cross-dataset generalization under both point and probabilistic forecasting metrics.

\subsection{In-distribution Forecasting}
We evaluate in-distribution forecasting on 29 Monash datasets~\cite{Monash}. Only training portions are included in LOTSA, while test sets are reserved for evaluation. Results are reported as normalized MAE relative to a naive forecast and aggregated by the geometric mean across datasets.
\begin{figure*}[!t]
\begin{center}
\centerline{\scalebox{1}[0.88]{\includegraphics[width=\textwidth]{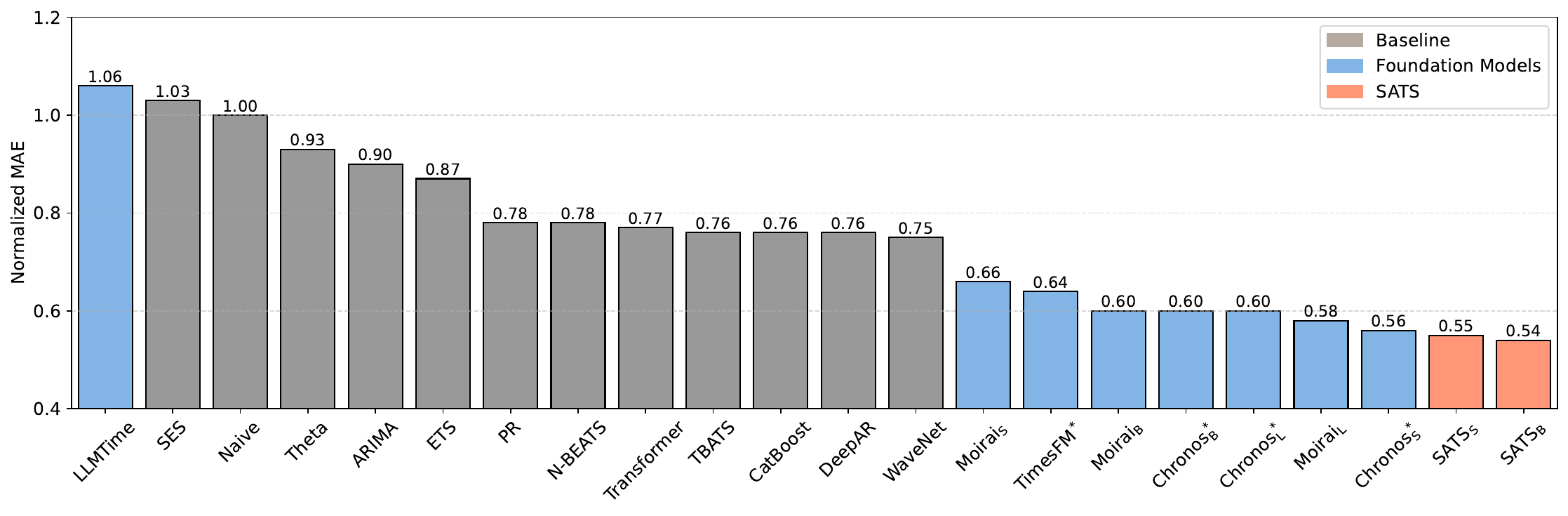}}}
\caption{In-distribution forecasting performance evaluated on 29 datasets from the Monash benchmark~\cite{Monash}. Methods trained with access to these evaluation datasets during pretraining are denoted with asterisks (*). Results are normalized using the naive forecast and summarized with the geometric mean.
}
\label{fig:in}
\end{center}
\end{figure*}

\begin{table*}[!t]
\begin{minipage}[t]{0.49\textwidth}
\centering
\caption{Module ablation under zero-shot evaluation. \boldres{Red} and \secondres{Blue} indicate the best and second-best results.}
\label{tab:ablMod}
\resizebox{\linewidth}{!}{%
\begin{tabular}{c|cccccc
>{\columncolor[HTML]{EFEFEF}}c
>{\columncolor[HTML]{EFEFEF}}c }
\hline
\textbf{Model variants} & \multicolumn{2}{c}{\textbf{ETTh}} & \multicolumn{2}{c}{\textbf{ETTm}} & \multicolumn{2}{c}{\textbf{Weather}} & \multicolumn{2}{c}{\cellcolor[HTML]{EFEFEF}\textbf{GIFT-Eval}} \\
Metrics & MSE & MAE & MSE & MAE & MSE & MAE & MASE & CRPS \\ \hline
$\text{SATS}_\text{B}$ & \boldres{0.365} & \boldres{0.396} & \boldres{0.306} & \boldres{0.341} & \boldres{0.239} & 0.283 & \boldres{0.802} & \boldres{0.550} \\
w/o SA & \secondres{0.370} & 0.404 & \secondres{0.311} & \secondres{0.344} & \secondres{0.244} & \secondres{0.281} & 0.866 & 0.607 \\
w/o CM & 0.398 & 0.409 & 0.317 & 0.351 & 0.259 & 0.293 & \secondres{0.819} & \secondres{0.569} \\
w/o RM & 0.381 & \secondres{0.400} & 0.324 & 0.354 & 0.251 & \boldres{0.269} & 0.833 & 0.583 \\ \hline
\end{tabular}%
}
\end{minipage}
\hfill
\begin{minipage}[t]{0.49\textwidth}
\centering
\caption{Alignment ablation under zero-shot evaluation. A dash denotes a removed objective; \boldres{Red} and \secondres{Blue} indicate the best and second-best results.}
\label{tab:ablAlign}
\resizebox{\linewidth}{!}{%
\begin{tabular}{cc|cccccc
>{\columncolor[HTML]{EFEFEF}}c
>{\columncolor[HTML]{EFEFEF}}c }
\hline
\multicolumn{2}{c|}{$\text{SATS}_\text{B}$} & \multicolumn{2}{c}{ETTh} & \multicolumn{2}{c}{ETTm} & \multicolumn{2}{c}{Weather} & \multicolumn{2}{c}{\cellcolor[HTML]{EFEFEF}\textbf{GIFT-Eval}} \\
Close & Far & MSE & MAE & MSE & MAE & MSE & MAE & MASE & CRPS \\ \hline
Mean & Max & \boldres{0.365} & \boldres{0.396} & \boldres{0.306} & \boldres{0.341} & \secondres{0.239} & \secondres{0.283} & \boldres{0.802} & \boldres{0.550} \\
Mean & - & 0.388 & 0.410 & 0.353 & 0.368 & 0.269 & 0.294 & 0.945 & 0.846 \\
Mean & Min & 0.386 & 0.414 & 0.332 & 0.361 & 0.248 & 0.287 & 0.892 & 0.637 \\
Mean & Random & 0.372 & 0.403 & \secondres{0.312} & 0.352 & 0.243 & 0.288 & 0.821 & \secondres{0.567} \\
Max & - & \secondres{0.368} & \secondres{0.398} & 0.316 & \secondres{0.348} & \boldres{0.237} & \boldres{0.277} & \secondres{0.817} & 0.573 \\ \hline
\end{tabular}%
}
\end{minipage}
\end{table*}

\textbf{Result.} 
As shown in Figure~\ref{fig:in}, \text{SATS} consistently outperforms all competing methods. Compared to \text{Moirai}$_\text{L}$, the best baseline trained on clean data, \text{SATS}$_\text{B}$ achieves a \textbf{6.9\%} improvement while using \textbf{only 22.6\%} of its parameters. Similarly, against \text{Chronos}$_\text{S}$, the strongest baseline under data contamination, \text{SATS}$_\text{S}$ achieves superior performance with just \textbf{30.4\%} of its parameter count. Notably, the gain from \text{SATS}$_\text{S}$ to \text{SATS}$_\text{B}$ is modest, likely because in-distribution forecasting involves limited temporal complexity, where increasing model size yields diminishing returns.

\subsection{Ablation Studies\protect\footnotemark}
\footnotetext{ETTh/ETTm are means of ETTh1/ETTh2 and ETTm1/ETTm2 after horizon averaging.}

\subsubsection{Module Design}

We begin by performing ablation studies on the components of \text{SATS}$_\text{B}$, with results summarized in Table~\ref{tab:ablMod}. On ETT, removing SA consistently degrades performance, while discarding either HM component causes larger drops in most cases; Weather shows a similar MSE pattern, with relatively small MAE differences. These results indicate that HM is central to long-term forecasting, where random and contiguous masking jointly encourage robust temporal representations, while SA provides complementary gains by regularizing cross-scale token spaces. On GIFT-Eval, removing SA yields the worst results, whereas removing individual HM components has a smaller impact. This contrast suggests that HM mainly strengthens long-horizon dependency learning on LSTF, while SA is more critical for cross-dataset generalization. Together with the t-SNE visualizations in Section~\ref{sec:mechAna}, these findings show that SA forms a stable, structured embedding space and that the two components provide complementary benefits.

\subsubsection{Alignment Mechanism}

The key design of SA is to align mean embeddings while separating maximal embeddings, thereby preserving both cross-scale consistency and scale-anchor diversity. We therefore vary the pooling choices in Table~\ref{tab:ablAlign} to examine whether these two roles are necessary. Removing the max-embedding repulsion term causes a clear performance drop, confirming the importance of scale-anchor separation. Replacing max pooling with min or random pooling generally yields degraded or less stable results, indicating that maximal embeddings better capture scale-specific information. Applying alignment only to maximal embeddings also remains suboptimal in most cases, suggesting that this constraint is too weak to prevent scale-wise anchor collapse. Overall, bringing mean embeddings closer while pushing maximal embeddings apart is a robust design choice, consistent with the theoretical analysis in Section~\ref{Sec-theory-minmax}.

\subsubsection{Transformer Types}

\begin{table}[!tbp]
\centering
\caption{Transformer-type ablation under zero-shot evaluation. Values are average MSE or MASE; ``-'' denotes an inapplicable setting, and \boldres{Red} marks the best result.}
\label{tab:trans}
\resizebox{\columnwidth}{!}{%
\begin{tabular}{c|ccc|ccc|ccc}
\hline
Transformer & \multicolumn{3}{c|}{Encoder-Only} & \multicolumn{3}{c|}{Decoder-Only} & \multicolumn{3}{c}{Encoder-Decoder} \\ \hline
Variants & RAW & +SA & +HM & RAW & +SA & +HM & RAW & +SA & +HM \\ \hline
ETTh & 0.389 & 0.383 & \boldres{0.370} & 0.382 & \boldres{0.376} & - & 0.392 & 0.376 & \boldres{0.368} \\
ETTm & 0.355 & 0.320 & \boldres{0.311} & 0.353 & \boldres{0.325} & - & 0.366 & 0.334 & \boldres{0.323} \\
Weather & 0.250 & 0.258 & \boldres{0.244} & \boldres{0.247} & \boldres{0.247} & - & 0.253 & 0.250 & \boldres{0.244} \\
\rowcolor[HTML]{EFEFEF} 
GIFT-Eval & 0.901 & \boldres{0.839} & 0.866 & 0.892 & \boldres{0.834} & - & 0.912 & \boldres{0.842} & 0.870 \\ \hline
\end{tabular}%
}
\end{table}

We further assess architectural generality by integrating the proposed modules into three Transformer variants, omitting HM for the decoder-only model because its pre-training objective is incompatible with hierarchical masking. As shown in Table~\ref{tab:trans}, SA and HM bring consistent gains across applicable architectures: HM mainly improves long-horizon forecasting, while SA benefits cross-dataset generalization. Under the same setting, the decoder-only model performs best, likely due to denser training signals~\cite{scaleLawsLLM}, though its incompatibility with HM limits further gains. By contrast, the larger encoder-decoder variant often trails the encoder-only model, possibly because additional cross-attention layers introduce optimization instability and gradient dilution~\cite{hong-lee-2025-variance}. Overall, these results suggest that SA and HM provide architecture-agnostic improvements.

\subsection{Model Analysis}\label{sec:modAna}
\subsubsection{Mechanism Analysis}\label{sec:mechAna}
\begin{figure}[!tbp]
\begin{center}
\centerline{\includegraphics[width=\columnwidth]{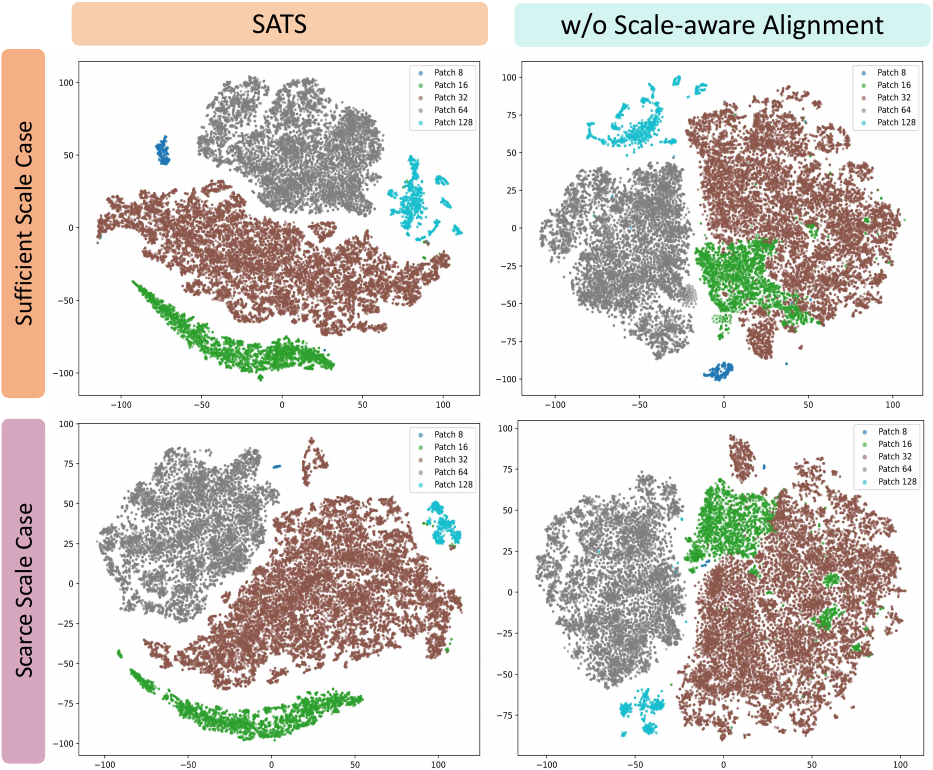}}
\caption{T-SNE token distributions in sufficient- and scarce-scale regimes. Colors denote patch-size origins.}
\label{fig:tsne}
\end{center}
\end{figure}

To understand what SA learns beyond forecasting gains, we examine whether it organizes multi-resolution tokens into a coherent yet scale-discriminative embedding space. We consider both a sufficient-scale regime with representative tokens at each scale and a scarce-scale regime with underrepresented patch sizes. In Figure~\ref{fig:tsne}, SATS forms compact yet separable clusters across patch sizes, even when size-8 tokens are scarce, whereas w/o SA exhibits confusion between patch sizes 16 and 32 and nearly absorbs scarce size-8 tokens. This visual pattern is consistent with Table~\ref{tab:quant}, where downsampled patches at resolutions \{8, 16, 32, 64\} are compared with the original size-128 embeddings. SATS achieves very high mean-pooling similarity (0.979--0.998), indicating cross-resolution alignment, while retaining lower max-pooling similarity (0.424--0.639), indicating preserved scale-specific anchors. Conversely, w/o SA shows weak or negative mean similarity (-0.054--0.293) but high max similarity (0.721--0.889), revealing fragmented global semantics and a tendency toward local-anchor collapse. Thus, the qualitative clusters and quantitative geometry tell a consistent story: SA pulls semantically matched resolutions toward a shared center while separating scale-specific maxima, producing coherent representations that ease Transformer pretraining and support generalization.
\FloatBarrier
\begin{table}[!tbp]
\centering
\caption{Cosine similarity between multi-resolution patch embeddings (\{8, 16, 32, 64\}) and size-128 representations under\\mean and max pooling.}
\label{tab:quant}
\resizebox{\columnwidth}{!}{%
\begin{tabular}{c|cccc|cccc}
\hline
\textbf{} & \multicolumn{4}{c|}{\textbf{Mean Pooling}} & \multicolumn{4}{c}{\textbf{Max Pooling}} \\ \hline
PatchSize & 8 & 16 & 32 & 64 & 8 & 16 & 32 & 64 \\ \hline
$\text{SATS}_\text{B}$ & 0.979 & 0.988 & 0.998 & 0.992 & 0.639 & 0.424 & 0.437 & 0.509 \\
w/o SA & 0.189 & -0.032 & -0.054 & 0.293 & 0.752 & 0.721 & 0.723 & 0.889 \\ \hline
\end{tabular}%
}
\end{table}

\subsubsection{Patch Protocol Sensitivity}

\begin{figure}[!tbp]
\centering
\vspace{5pt}
\centerline{\includegraphics[width=\columnwidth]{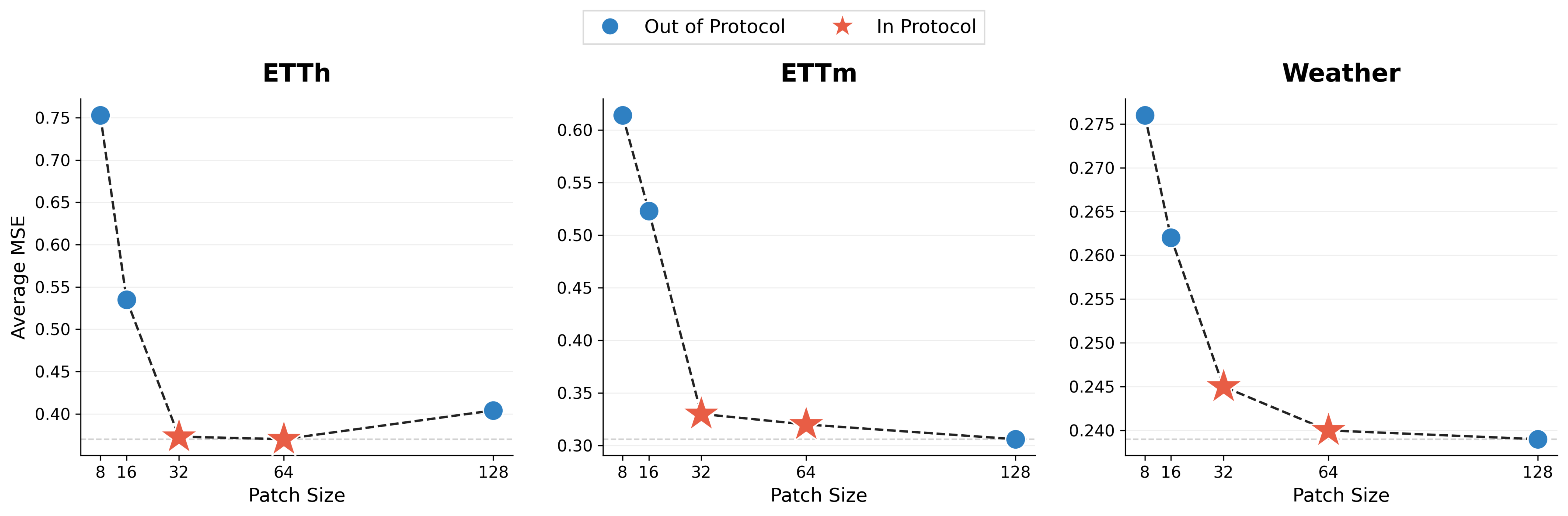}}
\caption{Sensitivity analysis of patch protocols on the LSTF benchmark. The charts report average MSE (lower is better) across different patch sizes. 
}
\label{fig:sen}
\vspace{10pt}
\centerline{\includegraphics[width=\columnwidth]{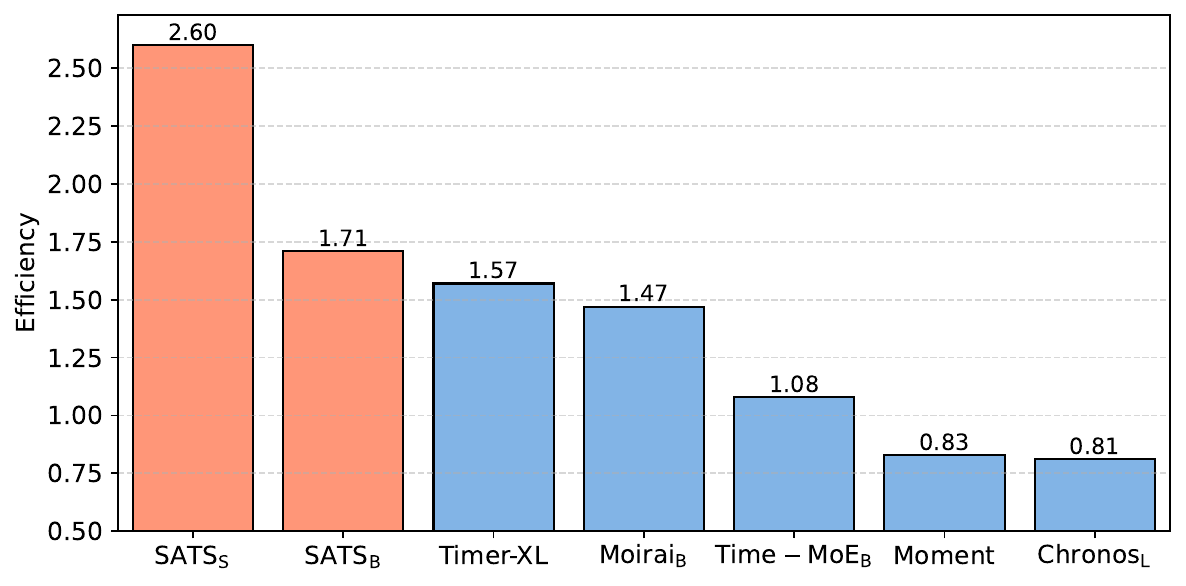}}
\caption{Model efficiency under zero-shot LSTF evaluation, measured as $\frac{1}{\text{MSE}\cdot \log(N + 1)}$. Higher values indicate better accuracy-parameter trade-offs for foundation models, whose parameter count meaningfully proxies capacity.}
\label{fig:eff}
\vspace{7pt}
\end{figure}
To further understand the mechanism of SATS and guide patch-size selection when sampling-frequency metadata is noisy or unavailable, we evaluate its sensitivity to patch protocols on the LSTF benchmark. As shown in Figure~\ref{fig:sen}, the prescribed protocol consistently yields the best performance, while deviations exhibit an asymmetric effect: using patch sizes smaller than the protocol causes substantial degradation, whereas using larger patch sizes leads to only mild performance loss. This suggests that excessively small patches fragment temporal semantics and fail to capture complete patterns~\cite{PatchTST}, while larger patches can still retain macroscopic trends through effective downsampling~\cite{MICN}. Therefore, in zero-shot scenarios with unreliable metadata, a conservative strategy is to prefer larger patch sizes within overlapping protocol ranges; when metadata is absent, spectral analysis can be used to select the closest frequency-based protocol.

\subsubsection{Model Efficiency}
To examine whether the gains of SATS translate into a practical accuracy-parameter trade-off, we evaluate model efficiency under the zero-shot LSTF setting. Since the proposed alignment and masking strategies introduce no additional learnable parameters to the multi-FFNs backbone, we measure efficiency as the inverse of the product between zero-shot error and the logarithm of model size. As shown in Figure~\ref{fig:eff}, SATS achieves a strong balance between accuracy and compactness: $\text{SATS}_\text{B}$ attains SOTA accuracy and surpasses the runner-up model, Timer-XL, by \textbf{8.9\%} in efficiency, while the lightweight $\text{SATS}_\text{S}$ delivers a \textbf{65.6\%} efficiency improvement despite only slightly outperforming Timer-XL in accuracy. These results indicate that SATS improves predictive performance without relying on parameter expansion, making it suitable for resource-constrained or real-time deployment.

\section{Conclusion}
This paper presents SATS, a \textbf{S}cale-\textbf{A}ware foundation model for \textbf{T}ime \textbf{S}eries that addresses the challenge of fragmented token spaces and misaligned representations in time series pretraining. SA is introduced to unify representations across patch sizes by jointly minimizing inter-scale embedding discrepancies and preserving scale-specific modeling capacity. Furthermore, HM combines RM and CM to capture temporal dependencies at multiple resolutions. Extensive experiments demonstrate that SATS achieves superior generalization and robustness while maintaining high efficiency, as its proposed alignment and masking strategies introduce no additional learnable parameters.

\bibliographystyle{IEEEtran}
\clearpage
\bibliography{ref}

\end{document}